\documentclass[a4paper,fleqn]{cas-dc}

\usepackage[numbers]{natbib}

\usepackage[ruled,linesnumbered,boxed]{algorithm2e}  

\def\tsc#1{\csdef{#1}{\textsc{\lowercase{#1}}\xspace}}
\tsc{WGM}
\tsc{QE}
\tsc{EP}
\tsc{PMS}
\tsc{BEC}
\tsc{DE}

\begin{document}
\let\WriteBookmarks\relax
\def\floatpagepagefraction{1}
\def\textpagefraction{.001}
\shorttitle{SCAI framework for Spectral data Classification}
\shortauthors{Yundong Sun et~al.}

\title [mode = title]{SCAI: A Spectral data Classification framework with Adaptive Inference for the IoT platform}                      



\author[1]{Yundong Sun}[type=editor,
                        orcid=0000-0001-7511-2910]

\credit{Conceptualization of this study, Methodology, Software}

\address[1]{School of Astronautics, Harbin Institute of Technology, Harbin, 150001, China.}

\author[2]{Dongjie Zhu}[style=chinese]
\cormark[1]

\author[1]{Haiwen Du}[%
   ]


\address[2]{School of Computer Science and Technology, Harbin Institute of Technology, Weihai, 264209, China.}

\author%
[2]
{Yansong Wang}
\author%
[1]
{Zhaoshuo Tian}

\cortext[cor1]{Corresponding author}


\begin{abstract}
Currently, it is a hot research topic to realize accurate, efficient, and real-time identification of massive spectral data with the help of deep learning and IoT technology. Deep neural networks have already played a key role in spectral analysis with their powerful capabilities of latent feature extraction and autonomous optimization learning for downstream tasks. Meanwhile, with the continuous innovation of deep learning computing and the increasing computing power of chips, more powerful models are constantly proposed and their performance on various tasks is constantly improved. However, the powerful performance of these models comes at the cost of a deeper and more complex structure. Moreover, the inference of most SOAT models is performed in a static manner, i.e., once the training is completed, the structure and parameters are fixed, and its computational load cannot be dynamically adjusted according to the computational capacity of the device, which severely limits its application. At the same time, not all samples need to allocate the same amount of computation to reach confident prediction, which causes a waste of resources and brings obstacles to maximizing the performance of the overall IoT platform. To address the above issues, in this paper, we propose a Spectral data Classification framework with Adaptive Inference (SCAI) for the IoT platform. Specifically, to allocate different computations for different spectral curve samples while better exploiting the collaborative computation capability among different devices under the IoT platform, we leverage Early-exit architecture, place intermediate classifiers at different depths of the architecture, and the model outputs the results and stops running when the prediction confidence at the current classifier position reaches a preset threshold. We propose a training paradigm of self-distillation learning, through which the output of the deepest classifier performs soft supervision on the previous classifier to maximize the performance and training speed of the whole architecture, especially the shallow classifier. At the same time, to mitigate the vulnerability of performance to the location and number settings of intermediate classifiers in the Early-exit paradigm, we propose a Position-Adaptive residual network (PA-ResNet). It can adjust the number of layers each block computes at different positions of the curve, so it can not only pay more attention to the information of important positions of the curve (e.g.: Raman peak), but also can accurately allocate the appropriate amount of computational budget based on task performance and computing resources, greatly alleviating the sensitivity of intermediate classifiers settings. To the best of our knowledge, this paper is the first attempt to conduct performance and efficiency optimization by adaptive inference computing architecture for the spectral detection problem under the IoT platform. We have conducted many experiments on liquors spectroscopy, and the experimental results show that our proposed method is not only well able to achieve adaptive inference under different computational budgets and different samples but also can achieve higher performance with less computational resources than existing methods.

\end{abstract}




\maketitle

\section{Introduction}

At present, the frequent occurrence of food safety incidents has seriously damaged the normal market operation order, violated the rights and interests of legitimate manufacturers and consumers, and even caused serious damage to the health of consumers. Among many food safety incidents, the cost of counterfeiting liquor is lower, and the price of genuine products is higher, so it has great economic benefits and has become the most frequent type of counterfeiting incident. How to achieve accurate and fast liquor authenticity detection has become an urgent problem to be solved \cite{tian2022development}. Spectral-based detection technology is widely used in food safety detection due to its advantages of simple operation, high sensitivity, low cost, and no pollution to the environment and samples, especially for the detection and analysis of liquors and other liquid food \cite{cozzolino2022advantages}.

With its powerful ability of latent feature extraction and autonomous optimization learning for downstream tasks, deep learning has played a key role in many fields, such as computer vision \cite{jiao2021new}, natural language processing  \cite{vaswani2017attention}, etc. At the same time, thanks to the continuous development of sensors, embedded devices, and communication protocol techniques, the Internet of Things (IoT) technology can make a large number of edge devices and cloud center computing devices work together and has been widely used in our production and life. For example, autonomous driving technology \cite{dong2022graph}, smart city \cite{jiang2021dl}, smart medical treatment  \cite{jan2021lightiot}, etc. At present, leveraging deep learning and IoT technology to achieve accurate, efficient, and massive real-time spectral detection is becoming a hot research direction.

In recent years, with the continuous innovation of deep learning and computing capabilities, more powerful deep learning models have been continuously proposed, and their performance on various tasks has been constantly refreshed \footnote{https://paperswithcode.com/sota/image-classification-on-imagenet}. However, the powerful performance of these models comes at the cost of a deeper and more complex structure. Moreover, the inference of most SOAT models is performed in a static manner, i.e., once the training is completed, the structure and parameters are fixed \cite{han2021dynamic}, and its computational load cannot be dynamically adjusted according to the computational capacity of the device. This severely limits its application, making it impossible for end devices in the IoT (such as smartphones, etc., which have low performance or require strict control of power consumption). At the same time, not all samples need to allocate the same amount of computation to reach confident prediction \cite{laskaridis2021adaptive}, which causes a waste of resources and brings obstacles to maximizing the performance of the overall IoT platform.

In big data scenarios and the IoT platforms, different devices have different computing capabilities and battery life, and at the same time, different application scenarios have different latency requirements. How to achieve high computing efficiency under the premise of ensuring high recognition accuracy, and how to allocate the corresponding computing load according to different device performances and samples is an urgent problem to be solved.

\begin{figure}
	\centering
	\includegraphics[width=3.2in,trim=0 0 0 0]{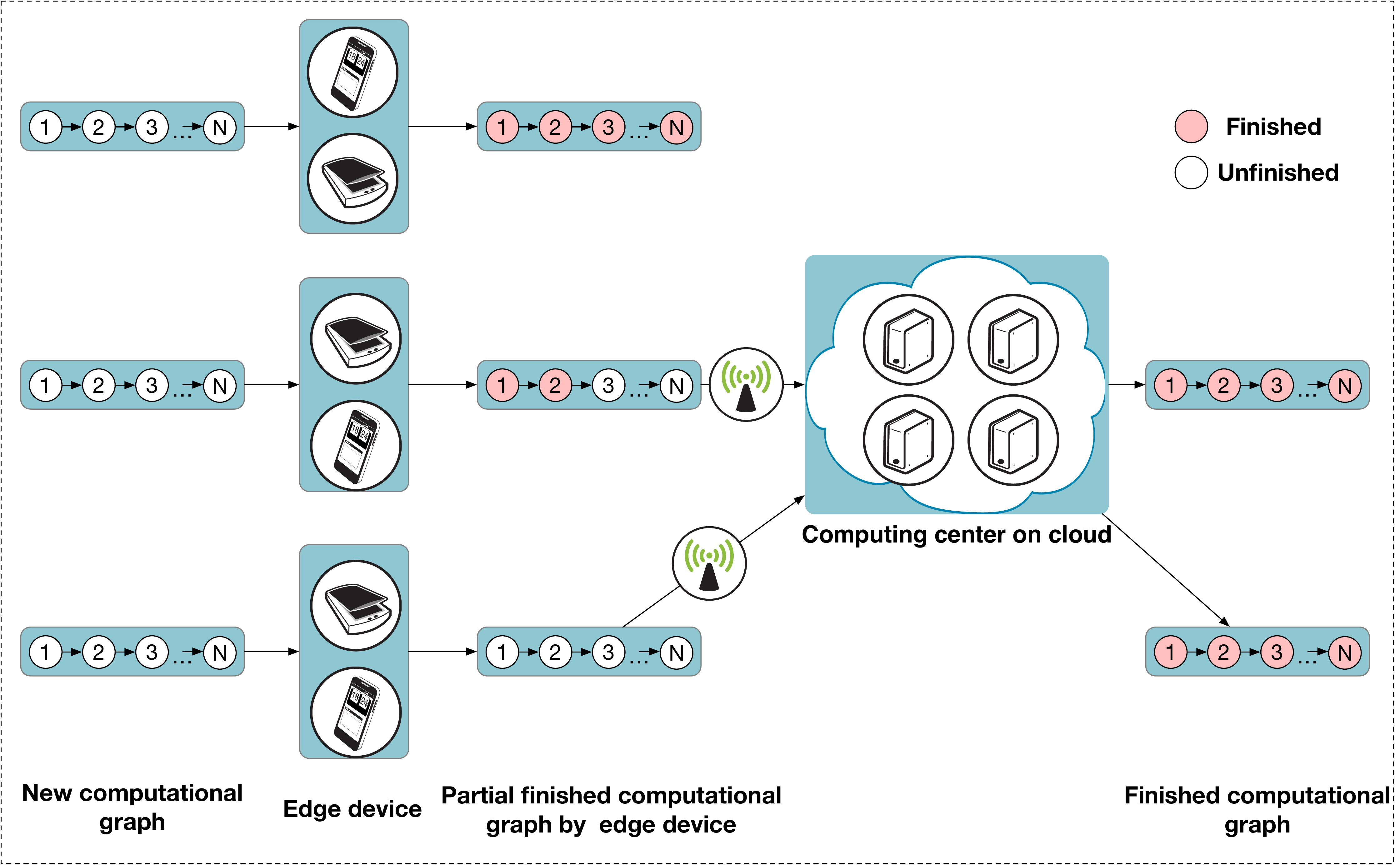}
	\caption{The schematic diagram of task offloading in IoT.}
	\label{FIG:1}
\end{figure}

Aiming at the above problems, this paper proposes a spectral data classification framework with elastic computing capabilities. Specifically, to allocate different computations for different spectral curve samples while better exploiting the collaborative computation capability among different devices under the IoT platform, we leverage Early-exit architecture, place intermediate classifiers at different depths of the architecture, and the model outputs the results and stops running when the prediction confidence at the current classifier position reaches a preset threshold. Meanwhile, the architecture can well support the task offloading of edge devices \cite{wang2021multi}. When the performance of the edge device does not reach the preset threshold, it can send the intermediate feature information to the high-performance device (such as the MEC server) for continued calculation (see Fig.\ref{FIG:1}). It can better exert the collaborative computing ability between various devices under the IoT platform.

Under the early-exit architecture, the most intuitive training strategy is to input the original features of the training samples into the model and combine the labels of the training samples and the predicted output of each classifier to calculate the loss (such as cross-entropy loss). Ideally, classifiers located in the deep layers will perform better than those located in the shallower layers. In such a case, the early classifier cannot achieve high accuracy as soon as possible without the knowledge guidance of the later classifier during the training phase. When we propose an adaptive inference architecture, we expect the model to have adaptive inference capabilities (the more computations, the better the performance), and at the same time, we also expect a shallower classifier with better performance. In other words, we prefer to get as much performance as possible with a small computational budget. We propose a training paradigm of self-distillation learning, through which the output of the deepest classifier performs soft supervision on the previous classifier to maximize the performance and training speed of the whole architecture, especially the shallow classifier.

Meanwhile, under the early-exit architecture paradigm, the overall performance of the model (including the performance of multiple intermediate classifiers and the final classifier) is easily affected by the location and number of intermediate classifiers. The decision of the location and number of intermediate classifiers not only affects the granularity of intermediate classifiers but also the overall overhead of early-exiting compared to the static inference \cite{laskaridis2021adaptive}. To solve this problem, we propose a Position-Adaptive residual network (PA-ResNet). It can adjust the number of computation layers in each block at different positions of the curve, so it can not only pay more attention to the information of important positions of the curve (e.g.: Raman peak), but also can accurately allocate the appropriate amount of computational budget based on task performance and computing resources, greatly alleviating the sensitivity of intermediate classifiers settings.

\section{Related Work}

\subsection{Spectral Classification Algorithms}

Initially, methods such as mass spectrometry, gas chromatography, gas chromatography-mass spectrometry, gravimetric methods, ultraviolet and visible spectrophotometry, infrared spectrophotometry, and fluorescence spectrophotometry  \cite{wang2006forensic, han2018development} were used to identify substances. However, these methods have poor reproducibility, relatively long analysis times, and high maintenance costs. Later, spectral-based detection technology is widely used in food safety detection due to its advantages of simple operation, high sensitivity, low cost, and no pollution to the environment and samples, especially for the detection and analysis of liquors and other liquid food \cite{cozzolino2022advantages}.

Among the existing spectral detection algorithms, the traditional matching algorithm based on curve characteristic peak matching still occupies a dominant position. This kind of methodology employs different similarity measurement algorithms, such as Euclidean distance and Pearson correlation coefficient, to obtain the matching score between the spectral curve of the test sample and the standard sample curve in the database, and then to judge the authenticity of the sample \cite{rousaki2021development}. In recent years, some studies have improved and optimized the existing methods for specific application problems. Tian et al. \cite{tian2022development} proposed a laser intensity calibration method based on the Raman peak of pure water, which can reduce the influence of different spectroscopic instruments, and the instability of laser power, and the change in the external environment. Simultaneously, before executing the Pearson correlation coefficient matching algorithm, a laser intensity selection algorithm is added to divide the laser intensity into 10 classes. Only the spectral data of the same or adjacent classes are selected for matching, which can reduce the number of comparisons with the spectral data in the database and can also avoid matching errors caused by similar curves of different substances under different classes of intensities, finally improving both the recognition speed and accuracy. Sun et al.\cite{sunyundong2022} found that in practical application scenarios, algorithms such as Pearson matching performed poorly in the discrimination between different substances due to the error of instrument laser intensity and system control. They innovatively proposed a spectral curve processing algorithm based on Raman peak alignment. Before the matching calculation, the sample curve and the target curve are numerically aligned according to the Raman peak, which can greatly alleviate the laser intensity emission errors of different instruments and control errors caused by control systems. They verified the proposed algorithm in the real liquor detection scenario. The experimental results show that the proposed algorithm can significantly improve the detection accuracy compared with the matching algorithm based on the Pearson correlation coefficient and has better discrimination accuracy and stability between different samples.

However, this type of algorithm cannot extract features and learn automatically and needs to manually extract, set, and adjust feature peaks based on domain knowledge, which cannot be effectively applied in the face of massive data. The machine learning model can realize complex feature extraction and analysis tasks without artificial feature extraction, and it has better generalization ability than artificially designed features. Li et al. \cite{li2018discrimination} used different statistical machine learning methods to classify and identify fresh pork fat, skin, ham, loin, and tenderloin muscle tissues. Both KNN and SVM classifiers could clearly distinguish fat, skin, and muscle tissues, with an accuracy of over 0.998, a sensitivity of over 0.995, and a specificity of over 0.998. However, almost all statistical machine learning methods require special preprocessing such as baseline correction and/or PCA as basic steps. Deep learning attempts to extract high-level features directly from the raw data, so the dependence on feature engineering can be further reduced, which is the main difference between deep learning and traditional machine learning algorithms. Liu et al.\cite{liu2017deep} proposed a convolutional neural network-based solution to realize chemical identification from raw spectral data without a preprocessing process. Yang et al.\cite{yang2018mine} proposed a novel water inrush water source discrimination model using laser-induced fluorescence (LIF) technology and convolutional neural network (CNN) to achieve online discrimination of mine water inrush sources, which can also reduce the involvement of manual data processing.

\subsection{Adaptive Inference}

In recent years, with the continuous development of deep learning, more powerful deep learning recognition models have been proposed, and recognition performance is getting better and better. However, the powerful performance of these models comes at the cost of a deeper and more complex structure. Moreover, the inference of most SOAT models is performed in a static manner, and its computational load cannot be dynamically adjusted according to different samples and the computational capacity of the devices. This severely limits its application in the IoT. To cope with the discrepancy in computing power and battery life of different devices under the IoT platform in big data scenarios, relevant scholars have devoted themselves to adaptive inference research.

Adaptive inference methods can be divided into three categories: sample-wise, spatial-wise, and temporal-wise. The sample-wise methods adaptively select the corresponding network structure and parameters according to different samples; the spatial-wise methods perform adaptive computation for features of different spatial regions or locations of the sample, and different computational budgets are put into at different moments of the sequence data by Temporal-wise methods  \cite{han2021dynamic}. The data types processed by these three kinds of methods are different, and the granularity of the corresponding processing units is also different. More closely related to this paper are the sample-wise and spatial-wise methods, so this section mainly reviews some research related to these two types of methods.

The intuition of the spatial-wise method is that not all input features are informative to the final prediction result. Spatial-wise methods can be divided into Point-wise, Region-wise, and Resolution-wise according to the granularity of processing units  \cite{han2021dynamic}. The point-wise method adjusts its structure or parameters for each feature point of the data (such as the pixel point of the image). Specifically, it can strategically only process some feature points \cite{verelst2020dynamic, xie2020spatially, li2021dynamic} or apply additional computation to some important feature points \cite{hua2019channel, kirillov2020pointrend}.  Region-wise methods perform parameterized transformations \cite{chen2021dynamic, kahatapitiya2021coarse, hu2021global, wang2020glance} or select the "class-discri-minative" regions through the Hard-attention mechanism \cite{cordonnier2021differentiable, rangrej2021probabilistic} to achieve higher accuracy or more efficient computation. The main steps of the Resolution-wise methods include: first, the entire feature map is scaled to different resolutions; second, the features with different resolutions are selectively processed through different branch networks in turn from small to large, so as to minimize the amount of computation while maintain ensuring accuracy \cite{yang2020resolution, wang2021not}.

\begin{figure*}
	\centering
	\includegraphics[width=7in,trim=0 0 0 0]{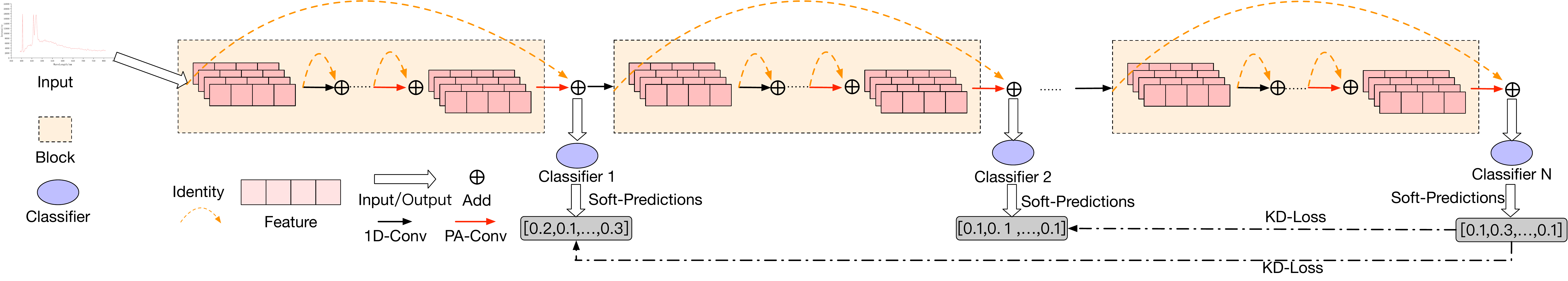}
	\caption{The schematic diagram of the overall architecture of SCAI.}
	\label{FIG:2}
\end{figure*}

Compared with the spatial-wise methods, the sample-wise methods regard the sample as the processing unit and have a coarser granularity. It can select different model structures (including dynamic model depth \cite{huang2017multi, bolukbasi2017adaptive, wang2018skipnet, liu2021faster, xia2021fully}, model width \cite{hou2020dynabert, ma2018modeling} and computation path \cite{bejnordi2019batch, liu2018dynamic}, etc.) according to the sample with different hardness, and can also adjust the model parameters. With the continuous expansion of the depth of deep learning models, dynamic model depth has also become the simplest, most direct, but most effective method among the adaptive inference methods. The dynamic model depth can adopt the early-exit strategy and the layer-skipping strategy. The early-exit strategy allows "simple" samples to output results at shallow exits without continuing to perform subsequent deep computations; the layer-skipping strategy selectively skips some layers according to the input. Unlike the early-exit strategy, which has multiple exits, it has only one exit at the end of the model. Layer-skipping strategies often require complex decision networks, such as reinforcement learning \cite{wang2018skipnet, wu2018blockdrop}, and thus, they need costly training and learning procedure. Relatively speaking, the early-exit strategy can achieve adaptive inference by setting different confidence thresholds without the use of a complex decision network, and the implementation, training, and learning costs are lower. Meanwhile, the early-exit strategy can well support the task offloading of edge devices \cite{wang2021multi}. When the performance of the edge device does not reach the preset threshold, it can send the intermediate feature information to the high-performance device (such as the MEC server) for continued calculation (see Fig.\ref{FIG:1}). It can better exert the collaborative computing ability between various devices under the IoT platform. Meanwhile, under the early-exit architecture paradigm, the overall performance of the model (including the performance of multiple intermediate classifiers and the final classifier) is easily affected by the location and number of intermediate classifiers. This decision of the location and number of intermediate classifiers not only affects the granularity of intermediate classifiers but also the overall overhead of early-exiting compared to the static inference \cite{laskaridis2021adaptive}. Although MSDNet \cite{huang2017multi} effectively alleviates this problem by introducing multi-scale and DensNet \cite{huang2019convolutional} structure, its complexity is high and it is not suitable for spectra data. Therefore, how to improve the early-exit strategy for spectral data is an urgent problem to be solved.

\section{Methodology}

Based on the previous analysis, we propose the SCAI architecture as shown in Fig.\ref{FIG:2}. Considering the complexity, the training difficulty of the model, and the support of task offloading in the IoT, we adopt the early-exits architecture. We propose a training paradigm of self-distillation learning, through which the output of the deepest classifier performs soft supervision on the previous classifier to maximize the performance and training speed of the whole architecture, especially the shallow classifier. This part of the content will be described in section \ref{sec31}. Further, in section \ref{sec32}, we propose a Position-Adaptive residual network (PA-ResNet), greatly alleviating the sensitivity of intermediate classifiers settings. We will detailly describe the deployment and application of this architecture on the IoT platform in section \ref{sec33}.

\subsection{SCAI: Early-exit with self-distillation}\label{sec31}
\subsubsection{Architecture}

As shown in Fig.\ref{FIG:2}, the overall architecture we propose consists of several blocks, and an intermediate classifier is inserted between every two blocks. In this paradigm, the early classifier will enforce the feature map in the early stage of the current model to be as beneficial as possible to the prediction results of the current classifier during the training phase, which is unbeneficial to the subsequent intermediate classifiers and final classifiers. Therefore, it will deteriorate the final performance of the overall architecture. To alleviate this problem, we propose a scheme of hierarchical ResNet. Specifically, we first adopt the ResNet structure inside each block, so that the block's original input information can be retained while learning higher-order features during the layer-by-layer forward process. Similarly, we also leverage the ResNet structure between different blocks, so that each classifier can obtain the shallow information at different distances or even the raw information while obtaining the high-level information learned by the previous blocks.

We denote the feature map output by the $s$-th layer network in the $l$-th block as $x_s^l$, and the feature map output by the last layer network in the $l$-th block as $x^l$, which is also the input of the $l$-th classifier. The original spectral input features are represented as $x_0^0$, so:

\begin{equation}
x_s^l=F_s^l(x_{s-1}^l)+x_{s-1}^l
\end{equation}

\begin{equation}
x^l=F^l (x^{l-1})+x^{l-1}
\end{equation}

\noindent where $F_s^l(\cdot)$ represents the function of the $s$-th layer network in the $l$-th block, $F^l(\cdot)$ is the function between the $l$-th block and the $l+1$-th block. In addition, to make the feature dimension and channel design between blocks more flexible, we add a feature transfer layer between every two blocks:

\begin{equation}
x_0^{l+1}=trans(x^l)=Conv1d(x^l)
\end{equation}

\noindent $x_0^{l+1}$ represents the input of the $l+1$-th block. For intermediate classifiers and final classifiers, we employ the simplest but most effective SoftMax classifier:

\begin{equation}
\hat{y_l}=softmax(x^l)
\end{equation}

\subsubsection{Training strategy}

Under the early-exit architecture, the most intuitive training strategy is to input the original features of the training samples into the model and combine the labels of the training samples and the predicted output of each classifier to calculate the loss (such as cross-entropy loss). Ideally, classifiers located in the deep layers will perform better than those located in the shallower layers. In such a case, the early classifier cannot achieve high accuracy as soon as possible without the knowledge guidance of the later classifier during the training phase. In this section, we propose a training paradigm of self-distillation learning, through which the output of the deepest classifier performs soft supervision on the previous classifier to maximize the performance and training speed of the whole architecture, especially the shallow classifier.

Specifically, there are two schemes: One scheme is the step-by-step distillation, that is, leveraging the soft prediction of the next classifier $\hat{y}_{s+1}$ to guide the current output $\hat{y}_{s}$. Another scheme is more intuitive, which directly guides the output of each intermediate classifier with the help of the output of the last classifier. After experimental verification, we find that the performance of the latter is better. Therefore, we adopt the latter scheme:

\begin{equation}
L_{task}^{(l)}=
\begin{cases}
   KL(\hat{y}_l,\hat{y}_{l+1}) &\text{if } l<L \\
   -\displaystyle\sum_{i=1}^C y_{i}log(\hat{y}_l) &\text{if } l=L
\end{cases}
\end{equation}

\noindent where $C$ represents the number of node labels, $y_i$ is the probability that the true label of the node is $i$, $y_i$ is 1 or 0.

\subsection{SCAI+: SCAI with PA-Resnet}\label{sec32}

\begin{figure}
	\centering
	\includegraphics[width=3.2in,trim=0 0 0 0]{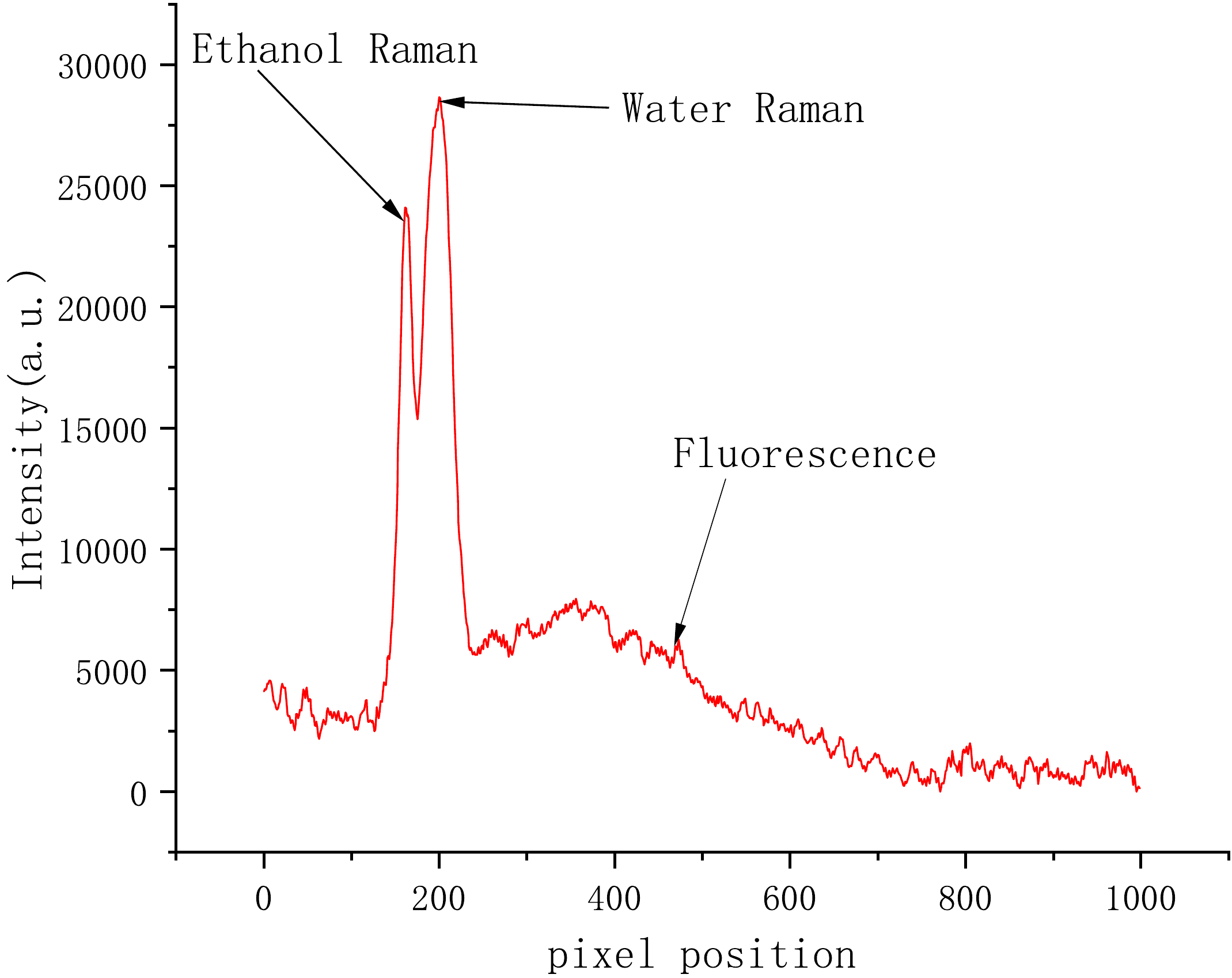}
	\caption{A typical liquor spectral curve.}
	\label{FIG:3}
\end{figure}

Although the early-exit paradigm has the advantages of low training cost and simple structure, the setting of the number and location of Intermediate classifiers has an important impact on its performance. Therefore, the determination of these two settings is a headache work. Meanwhile, similar to the fact that informative features may only exist in some regions of an image in computer vision \cite{han2021dynamic, figurnov2017spatially}. In the spectral curve (as shown in Fig.\ref{FIG:3}), except for the Raman peaks and fluorescence peaks of different substances with the most informative features \cite{tian2022development}, the other parts are not "class-discriminative" regions, and there are even many redundant features that will deteriorate the final result. Therefore, we believe that informative features may only exist in certain positions in the spectral curve.

Based on the above analysis, in this section, we propose a Position-Adaptive residual network (PA-ResNet). It can adjust the number of computation layers in each block at different positions of the curve, so it can not only pay more attention to the information of important positions of the curve (e.g.: Raman peak), but also can accurately allocate the appropriate amount of computational budget based on task performance and computing resources, greatly alleviating the sensitivity of intermediate classifiers settings. In this paper, we denote the enhanced SCAI with PA-Resnet as SCAI+.

\begin{figure}
	\centering
	\includegraphics[width=3.2in,trim=0 0 0 0]{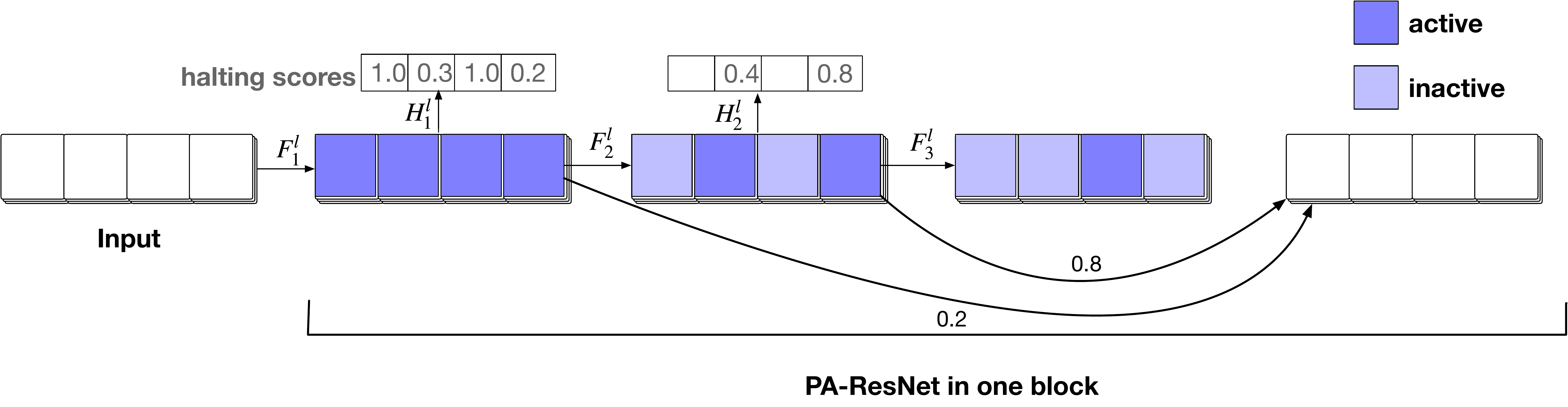}
	\caption{The schematic diagram of the PA-ResNet model structure within a block.}
	\label{FIG:4}
\end{figure}

Fig.\ref{FIG:4} shows the schematic diagram of the PA-ResNet model structure within a block. Inspired by ACT \cite{graves2016adaptive}, we add the prediction branch of the halting score based on the Residual unit. But our purpose is different from that of ACT. ACT wants to control where the inference of the model should stop through the halting score. It is oriented to the sample sequence, and the calculation of the halting score is oriented to each sample. In this paper, our purpose is to automatically choose to allocate different amounts of computation in a block at different positions of the curve, expecting to allocate more computation in more "class-discriminative" positions, while avoiding too much computation in indiscriminative positions. The SACT structure \cite{figurnov2017spatially} in the field of computer vision adopts a similar processing structure to accelerate the computational speed of image processing, its core purpose is to selectively allocate different amounts of computation in different regions of the image, which is similar to this paper. But SACT is not an early-stop architecture and alleviating the sensitivity of intermediate classifiers settings in the early-stop architecture is the core problem in this paper, which is the first attempt in the field.

\begin{figure*}
	\centering
	\includegraphics[width=7in]{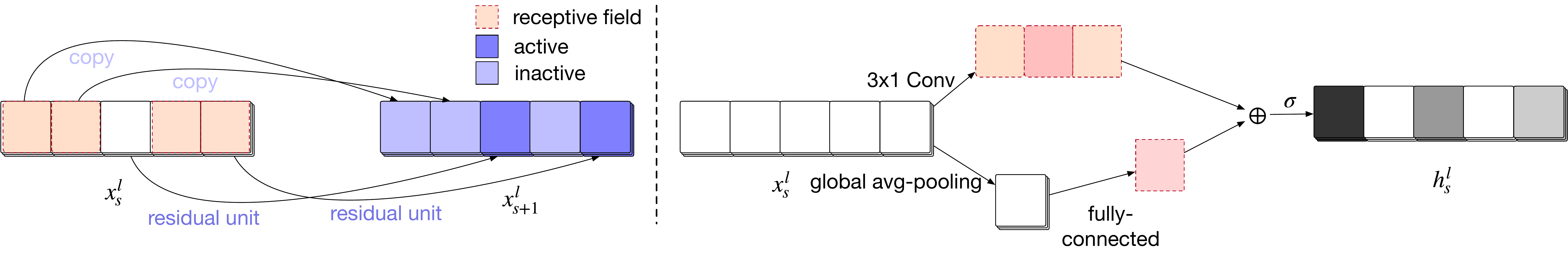}
	\caption{(a) is the Residual unit structure of PA-ResNet. The feature in the activated state will be used to extract higher-order features in the next Residual unit, and the feature in the inactive state is directly copied to the next layer without any operation. (b) is the calculation of the halting score, which is obtained through the fusion of local information and global information.}
	\label{FIG:5}
\end{figure*}

The Residual unit structure of PA-ResNet is shown in Fig.\ref{FIG:5} (a). We define the position of the feature map of the spectral curve as active or inactive in each layer of the network. The halting score of the feature map in the active position does not reach the preset threshold $\varepsilon$, which means that the feature in this position needs to be used to extract higher-order features in the next Residual unit; the halting score of the feature in the inactive position reaches the preset threshold $\varepsilon$, which means that the feature in this position is sufficient to meet the classification requirements of the intermediate classifier in this block. If the extraction of high-order features is continued, the classification performance will be weakened, the computational performance will be wasted, and the inference latency will be increased. Therefore, for inactive information, only a copy operation is required.

The calculation of the halting score is shown in Fig.\ref{FIG:5} (b). We believe that whether the feature information of the current position of the curve needs to be stopped depends not only on the current position and its local information but also on the global information of the whole curve. Therefore, we design a convolutional neural network to capture the current position and its local information, while leveraging global avg-pooling and fully connected layers to capture the global information of the whole curve. Finally, we use two kinds of information to calculate the halting score. Specifically, it can be expressed by the following equation:

\begin{equation}
H_{s}^{l}(x)=\sigma \big( W_s^l*x+ \widetilde{W_s^l}pool(x)+b_s^l \big)
\end{equation}

\noindent where $*$ represents a $3*1$ one-dimensional convolutional neural network with 1 channel, pool is global avg-pooling, $W_s^l$, $\widetilde{W_s^l}$ and $b_s^l$ are all learnable parameter matrices, $\sigma(x)=\frac{1}{1+exp(-x)}$. In particular, we force the halting score of the last layer in each block to 1:

\begin{equation}
h_{s}^{*}=H_s^*(x_s^*),s=1\dots (S-1)
\end{equation}

\begin{equation}
h_{S}^{*}=1
\end{equation}

For the information of a certain position in the active state, the number of computational layers in the inference process can be obtained by the following equation:

\begin{equation}
N=\min \bigg\{ n\in \{ 1\dots S\}:\displaystyle\sum_{s=1}^n h_s \ge 1-\varepsilon \bigg\}
\end{equation}

Referring to SACT \cite{figurnov2017spatially}, we also define retainer $R$:

Further, the distribution of the halting score is given by:

\begin{equation}
p_{s}=
\begin{cases}
   h_s &\text{if } s<N \\
   R &\text{if } s=N \\
   0 &\text{if } s>N
\end{cases}
\end{equation}

Finally, the final output feature of each block can be given by the weighted summation of the halting score:

\begin{equation}
output=\displaystyle\sum_{s=1}^{S} p_s x_s
\end{equation}

The overall calculation process is shown in Algorithm \ref{alg:1}.

\begin{equation}
R=1-\displaystyle\sum_{s=1}^{N-1} h_s
\end{equation}

\begin{algorithm}
\caption{Positionally Adaptive Residual network for one block.}
\label{alg:1}
\SetKwInOut{Input}{Input}
\SetKwInOut{Output}{Output}

\Input{The feature of the curve $x$ with shape $1*W$}
\Input{The residual units number of this block $S$}
\Input{Halting score threshold $0<\varepsilon<1$}
\Output{Output feature map of this block with shape $W\times C$}
\Output{ponder cost $\rho$}

$\hat x=input$\;
\For{$i=1$ \KwTo $len(\hat x)$}
{
	$\alpha_i=true$\;
	$c_i=0$\;
	$R_i=0$\;
	$output_i=0$\;
	$\rho_i=0$\;
}

\For{$s=1$ \KwTo $S$}
{

	\If {$\mathbf{not}$ $\alpha_i,\forall i \in \{ 1,2,\dots,len(\hat x)\}$}
	{
		$\mathbf{break}$\;
	}

	\For{$i=1$ \KwTo $len(\hat x)$}
	{
			\eIf {$\alpha_i$}
			{
				$x_i=F_s^l(\hat x)_i$\;
			}
			{
				$x_i=\hat{x_i}$\;
			}
	}

	\For{$i=1$ \KwTo $len(\hat x)$}
	{
			\If {$\mathbf{not}$ $\alpha_i$}
			{
				$\mathbf{continue}$\;
			}
			\eIf {$s<S$}
			{
				$h_i=H_s^l(x)_i$\;
			}
			{
				$h_i=1$\;
			}
			$c_i+=h_i$\;
			$\rho_i+=1$\;
			\eIf {$c_i<1-\varepsilon$}
			{
				$output_i+=h_i\cdot x_i$\;
				$R_i-=h_i$\;
			}
			{
				$output_i+=R_i\cdot x_i$\;
				$\rho_i+=R_i$\;
				$\alpha_i=false$\;
			}
	}
	$\hat x=x$\;
}
$\mathbf{return}\enspace output,\rho$
\end{algorithm}

At the same time, we also expect that while maintaining the performance of the model, the amount of computation is as small as possible, that is, the smaller the ponder cost, the better. Therefore, we define the loss function of the model's computational cost $L_cost$ and add it to the model training loss:

\begin{equation}
L=L_{task}+\gamma L_{cost}=\displaystyle\sum_{l=1}^{L} L_{task}^{(l)}+\gamma \displaystyle\sum_{l=1}^{L} \rho^{l}
\end{equation}

\noindent where $\gamma \ge 0$ is a regularization coefficient that controls the trade-off between optimizing the original loss function and the ponder cost. $\rho^l$ is the ponder cost of the $l$-th block.

\subsection{Deployment and Application in the IoT Platform}\label{sec33}

\begin{figure}[htbp]
	\centering
	\includegraphics[width=\linewidth,scale=1.00]{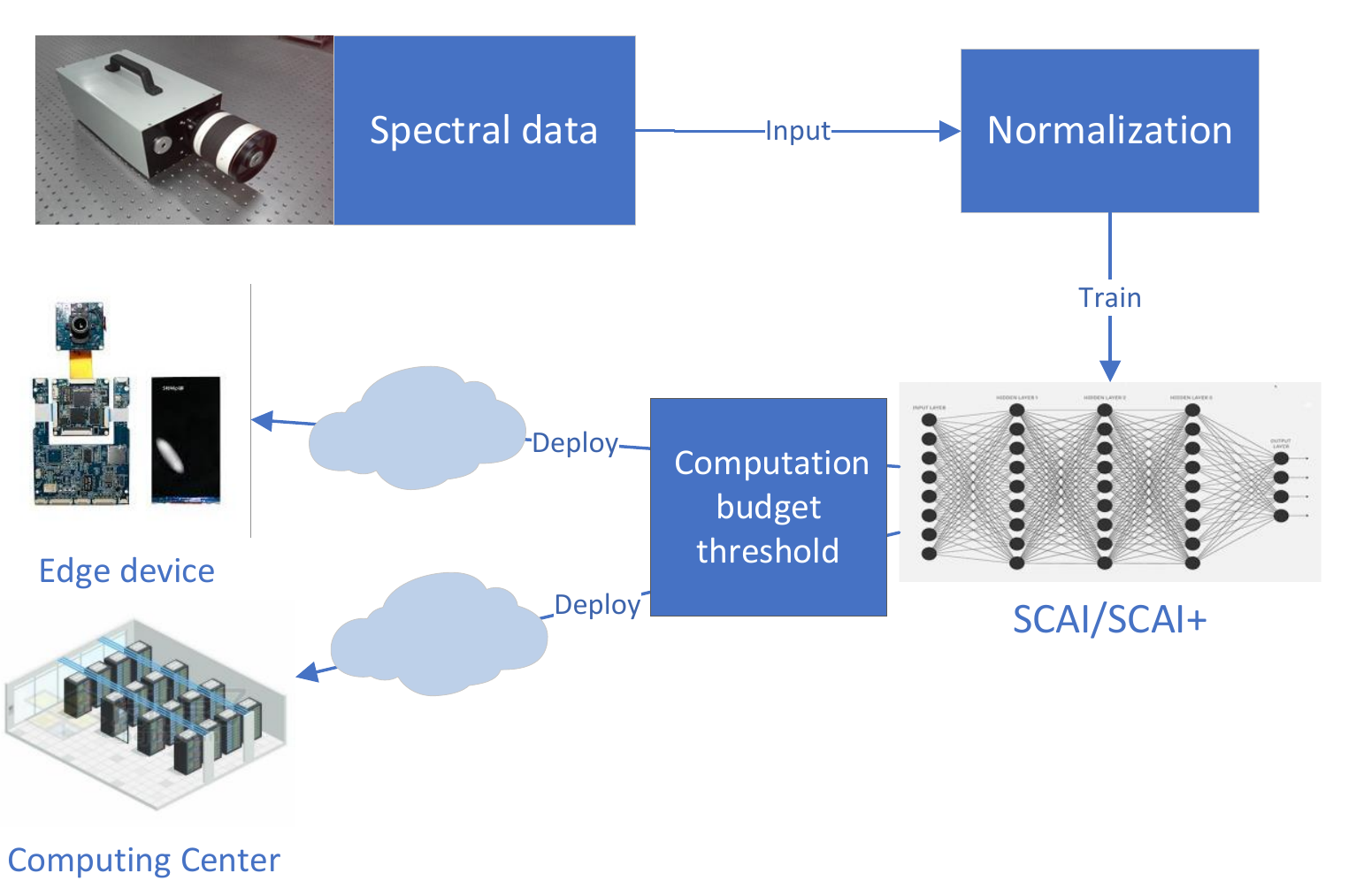}
	\caption{The schematic diagram of training and deployment scheme of the SCAI model in the IoT.}
	\label{FIG:6}
\end{figure}

The schematic diagram of the training and deployment scheme of the SCAI model in the IoT is shown in Fig.\ref{FIG:6}. First, we use the spectral signal collector to obtain a large amount of sample data. After the process of data normalization, the model is trained to obtain a spectral classification model that meets the expected performance. The above processes are completed in the laboratory. Finally, after setting the corresponding maximum computing budget $B_{max}$ according to the performance of different devices, we then deploy the trained models to different edge devices and computing centers respectively.

\subsubsection{Inference strategy}\label{sec331}

According to different application scenarios, the inference process can be divided into Anytime inference with a single sample and Budgeted batch inference with batch samples \cite{huang2017multi}. We will introduce them separately.

\begin{itemize}
\item {Anytime inference}
\end{itemize}

\begin{figure}[htbp]
	\centering
	\includegraphics[width=\linewidth,scale=1.00]{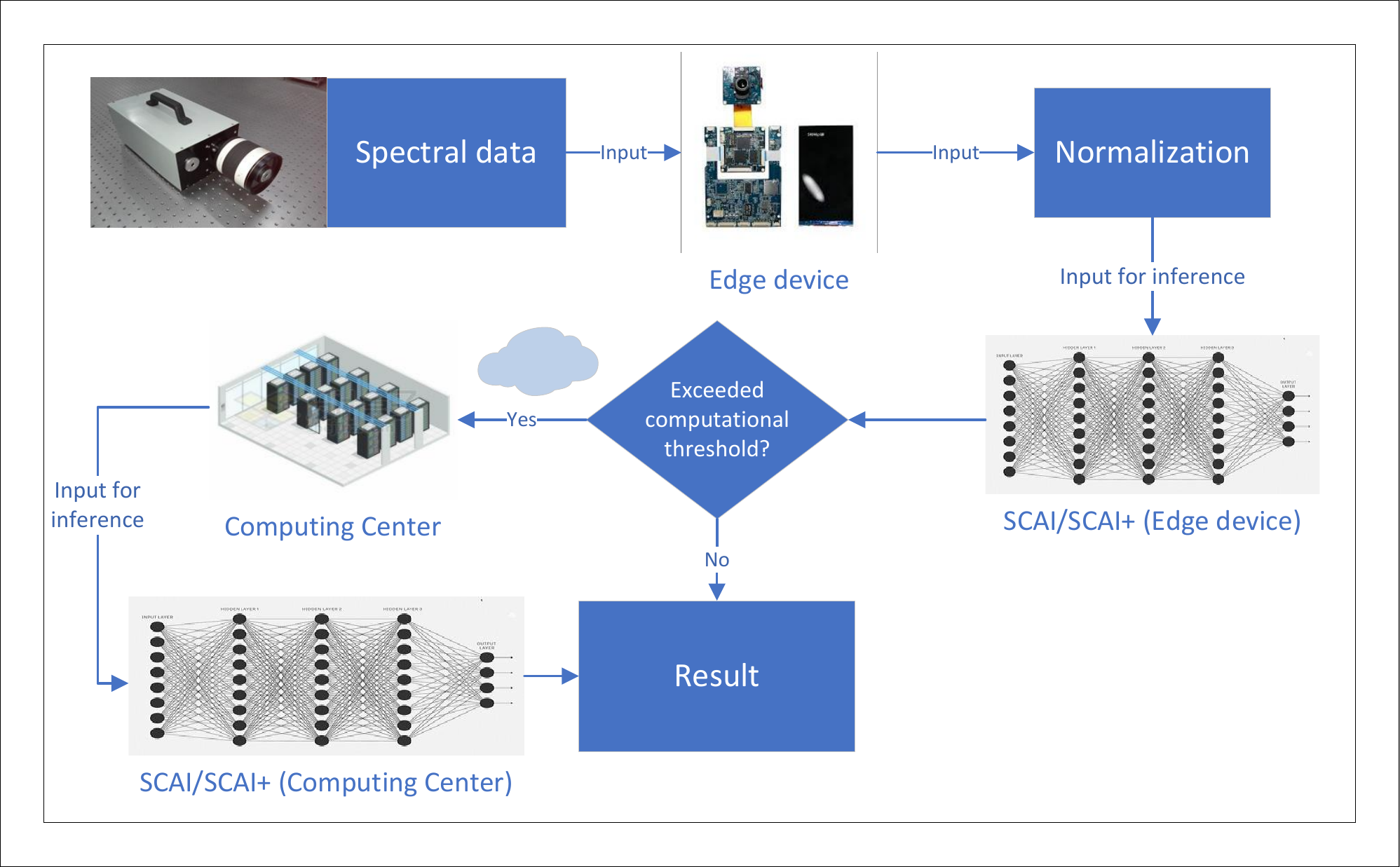}
	\caption{The schematic diagram of the application scheme of anytime prediction in the IoT.}
	\label{FIG:7}
\end{figure}

For the most common scenario of single-sample rapid prediction, it means that the neural network model needs give the prediction result of a single spectral sample with minimal latency. For the most common scenario of single-sample rapid prediction, it means that the neural network model needs give the prediction result of a single spectral sample with minimal latency. This procedure is usually done by different edge devices, and it is certainly impossible to train different neural network models for different devices. The adaptive inference computing framework proposed in this paper can be deployed to different devices after one training and maximize the performance on all edge devices. In the application of anytime inference, we propagate the input through the network until the budget is exhausted and output the latest prediction. More importantly, the model proposed in this paper can combine end devices and high-performance computing centers to maximize the computing performance of the entire IoT platform. The schematic diagram of the application scheme of anytime prediction in the IoT is shown in Fig.\ref{FIG:7}. For the scenario of anytime prediction, each test sample has a finite computational budget B in the prediction phase. The computational budgets are non-deterministic and vary for each test instance. It is determined by the occurrence of events that require the model to output predictions immediately. We assume that the budget comes from some joint distribution of $P(x,B)$, and in some applications, $P(B)$ may be independent of $P(x)$ and can be estimated. For example, if the event is governed by a Poisson process, then $P(B)$ is an exponential distribution.

For the edge device, if it generates a prediction of sample x within the budget $B$ and satisfies $B<B_{max}$, the prediction is performed on this device, and the result can be fed back directly; if $B>B_{max}$, it means that the computational budget of this sample exceeds its budget threshold or the latency is not acceptable for this device, and this prediction is not performed on this device (or only part on this device, similar to the task offloading in Fig.\ref{FIG:1}), the original data or the intermediate features are passed to the computing center for subsequent analysis and prediction. $B_{max}$ is the corresponding maximum computational budget preset according to the performance of the specific device when the model is deployed. We denote the loss of a model $f(x)$ as $L(f(x),B)$, $f(x)$ must produce a prediction for a sample $x$ within budget $B$. The goal of the model at any time is to minimize the expected loss under the budget distribution: $L(f)\cdot E[L(f),B]_{P(x,B)}$. Here, $L(\cdot)$ denotes a suitable loss function. The simplest way is that the expectation of $P(x,B)$ can be estimated by the mean of the samples of $P(x)$ on the validation dataset.

\begin{itemize}
\item {Budgeted batch inference}
\end{itemize}

The budgeted batch inference is mainly executed on a high-performance computing center, which means that many samples in a batch share a fixed computational budget. The model can adaptively allocate less computation to "simple samples" and more computation to "hard samples" in this batch. For example, there is a large amount of spectral data we need to process, if we can save some computation on "simple samples", it is more efficient for the overall computational budget. In such a batch, if the total computational budget is limited to B, and this batch contains $M$ samples, the computation allocated to "simple samples" should be less than $B/M$, and to the "hard samples" should be more than $B/M$.

In the budgeted batch inference, the model needs to classify the sample set $D_{test}\cdot\{x_{1},x_{2},\dots,x_{M}\}$ within a finite computational budget $B>0$ known in advance. The learner's goal is to minimize the prediction loss for all examples in $D_{test}$, within a computational budget that does not exceed $B$, which can be denoted by $L(f(D_{test}),B)$.

How to set the exit threshold $\theta_l$ of each classifier is a key problem in early-exit architecture. In this paper, we refer to the strategy of MSDNet \cite{huang2017multi}. Specifically, for a sample, if its prediction confidence (we regard the largest output probability among all output of SoftMax as the confidence.) output by classifier $l$ exceeds a threshold $\theta_l$, then the inference of this sample will exit after the classifier $l$ and this result will be the final result. Before training, we calculate the computational budget $C_l$ required by the network to perform to the $l$-th classifier. We denote a fixed dropout probability by $0<q\le 1$, that is, samples arriving at the classifier can obtain a classification result and exit with sufficient confidence. Before training, we calculate the computational budget $C_l$ that the network exits after the $l$-th classifier. We assume $q$ to be consistent across different classifiers. Based on this, we can calculate the probability that a sample exits after the $l$-th classifier $q_l=z(1-q)^{(l-1)} q$, where $z$ is a normalization constant that ensures $\sum_l q_l=1$. During the test phase, we need to ensure that the total computational budget for classifying all samples in the $D_test$ does not exceed $B$ (as expected). This produces the constraint $|D_{test}|q_{l} C_l\le B$. We can approximate $|D_{test}|q_{l}$ on the validation set $D_{valid}$ with the value of the samples exiting after the $l$-th classifier, and then solve this constraint of $q$ and determine the threshold $\theta_l$ on the $D_{test}$.

\section{Experiment}

To verify the performance of the SCAI architecture proposed in this paper, in this section, we use our self-developed micro-fluorescence spectrometer to collect a large number of real liquor spectral data in the market and conduct a lot of experimental work based on this. First, we conduct comparative experiments (Anytime Prediction and Batched Budget Prediction) to verify the superiority of our method and the significant improvement of the proposed core modules, in comparison with current state-of-the-art methods and variant models, respectively. In addition, to further analyze the advantages brought by PA-Resnet and the self-distillation training paradigm, we conducted relevant exploration experiments respectively.

\subsection{Experiment Instrument}

\begin{figure}[t]
	\centering
	\includegraphics[width=\linewidth]{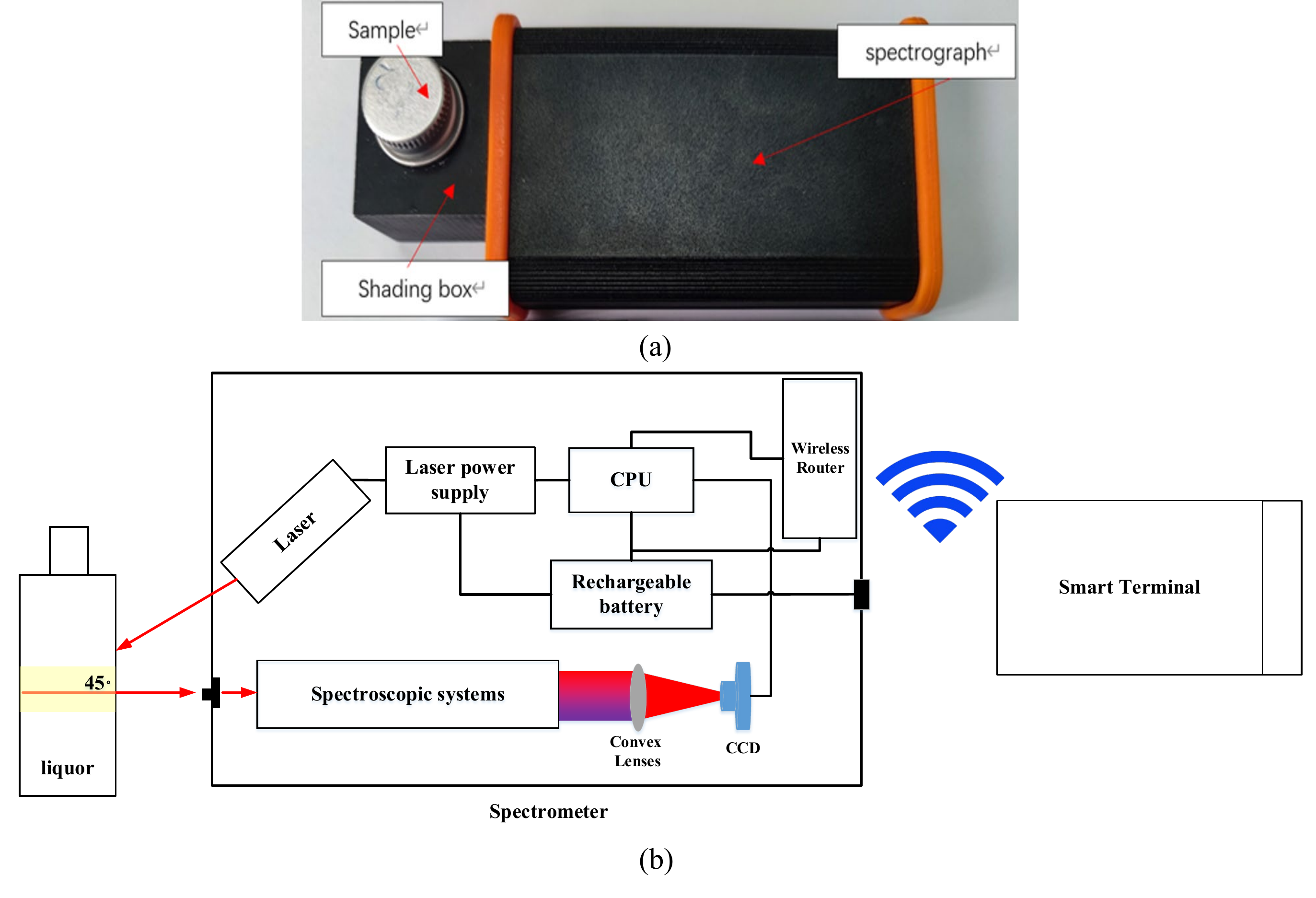}
	\caption{(a) The physical diagram and (b) the internal structure of the experimental system.}
	\label{FIG:8}
\end{figure}

The physical diagram of the experimental system is shown in Fig.\ref{FIG:8} (a). The size of our self-developed micro-spectrometer is 103mm*58mm*25mm. The laser diode has a wavelength of 405 nm and can be set to continuous or pulse output mode as required. The CCD pixel is 1280*720, and the CPU chip is Atheros AR9331 SOC. Fig.\ref{FIG:8} (b) shows the internal structure of the experimental system. The system includes a laser spectrometer and a smart terminal. The laser spectrometer includes a CPU module, a spectroscopic system, convex lenses, and a battery. The CPU module is respectively connected with the laser emission module, the CCD, and the wireless routing module. The laser emitted from the laser emission module excites the tested liquor through the wall of the sample bottle, and the optical signal generated by the spectroscopic system is then concentrated on the surface of the CCD by the convex lens. The CCD is used to convert the optical signal into an electrical signal, the CPU module is used to receive the control instructions of the smart terminal and control the laser emission module to emit laser with preset laser intensity, and convert the electrical signal transmitted by the CCD into a digital image. The wireless router module is used to realize wireless communication between the laser spectrometer and the smart terminal. When conducting experiments, we put liquor samples in a clear glass bottle, which were placed into the shading box for testing. The spectral data is transmitted to the smart terminal through the wireless router.

\subsection{Dataset}

The tested liquors were selected from 12 brands in the Chinese market, including 3 Maotai-Flavor liquors, 3 Luzhou-Flavor liquors, and 6 Fen-Flavor liquors. The names and numbers are Beijing Erguotou (\# 1), Weihai Wei Shao Guo (\# 2), clove love (\# 3), the raw pulp (\# 4), Beidacang (\# 5), bitter mustard (\# 6); Jing Zhi Bai Gan (\# 7), Hengshui Lao Bai Gan (\# 8), Lao Jiu Hu (\# 9), Niu Lan Shan aged (\# 10), du Er Jiu (\# 11) and Xiao Lang Jiu (\# 12). 100 sample data were taken at different times for each liquor. In the experiment, we split the data according to train: validation: test=8:1:1. To ensure the accuracy and consistency of the measurement results, we must first calibrate the laser intensity of the instrument, and the sample used for calibration is pure water. All obtained data are normalized and then input into different models for experiments.

\subsection{Experimental setup}

The parameters of the SCAI model proposed in this paper are initialized by random initialization, and the Adam optimizer \cite{huang2017multi} is used to optimize the model. The learning rate of Adam is set to 0.001, we adopt the early-stop strategy in the training process, and the patience is 50, i.e., we stop training if the validation performance does not increase for patience consecutive 50 epochs. For other hyperparameters, we leverage the grid search strategy to determine the optimal values. For example, the number of classifiers of the model, that is, the number of Blocks $L$, is searched in the range $\{1, 2, 3, 4, 5, 6\}$, and the optimal value finally is determined as 4. The number of model layers in each block, that is, the number of residual units S, is searched in the range of $\{1, 2, 3, 4, 5, 6, 7, 8\}$ to find the best combination. Considering the specific performance, the best combination is finally determined as $\{4, 4, 4, 4\}$, that is, the $S$ of each block is 4. We search for $\gamma$, the trade-off between optimizing the original loss function and the ponder cost, in $\{1e-6, 1e-5, 1e-4, 1e-3, 1e-2\}$ and find that the optimal value is  $\gamma=1e-5$. And the halting score threshold $\epsilon$ is searched in the range $\{0.01, 0.02, 0.05, 0.1, 0.2\}$, and the optimal value is finally determined as $\epsilon=0.02$. To ensure the stability of the experimental results, all the results in this paper are the average of 10 experimental results.

\subsection{Baselines}

\begin{itemize}
\item 
ResNet-1D \cite{he2016deep}(varying depth): Our method is proposed based on the ResNet network structure, so we first compare it with the ResNet network. Since the native ResNet is a static network structure, it cannot support adaptive inference. Therefore, we obtain networks with different computing budgets by changing the number of layers of the network to verify the performance of the model under different computing budgets. It is worth noting that the Resnet convolution unit in this paper is 1-dimensional.

\item 
DRSN-CS  \cite{zhao2019deep}(varying depth): DRSN is a recent work, which is also a variant of ResNet. This work inserts soft thresholding as a nonlinear transform layer into deep network structures to eliminate noise features. Its original intention is to eliminate noise in signal data, which is also suitable for the spectral curves. DRSN-CS is one of the schemes of DRSN, which has a building unit of channel sharing threshold. Similar to ResNet-1D \cite{he2016deep}(varying depth), we can also obtain networks with different computing budgets by changing the number of layers of the network to verify the performance of the model under different computing budgets.

\item
DRSN-CW \cite{zhao2019deep}(varying depth): DRSN-CW is another scheme of DRSN. Unlike DRSN-CS, a separate threshold is applied to each channel of the feature map.

\item
SCAI: The early-stop architecture proposed in section \ref{sec31} in this paper, which lacks the PA-Resnet network structure compared to SCAI+. We compare it with ResNet-1D, DRSN-CS, and DRSN-CW to explore \\ whether the early-stop architecture and self-distillation learning paradigm proposed in this paper have advantages compared with these methods. At the same time, we can find the performance improvement effect of the PA-Resnet network structure by combining it with SCAI+.

\item
SCAI+: The SCAI architecture with the PA-Resnet network structure, it is the complete body of the method in this paper. In theory, it should have the strongest performance and computational budget performance.

\item
$SCAI+_{w/o KD}$: We removed the self-distillation training paradigm on the basis of SCAI+ and obtained \\ $SCAI+_{w/o KD}$. Comparing it with SCAI+, we can see the superiority of the self-distillation training \\ paradigm.
\end{itemize}

\subsection{Comparative Results}

\begin{figure*}
	\centering
	\includegraphics[width=6.8in]{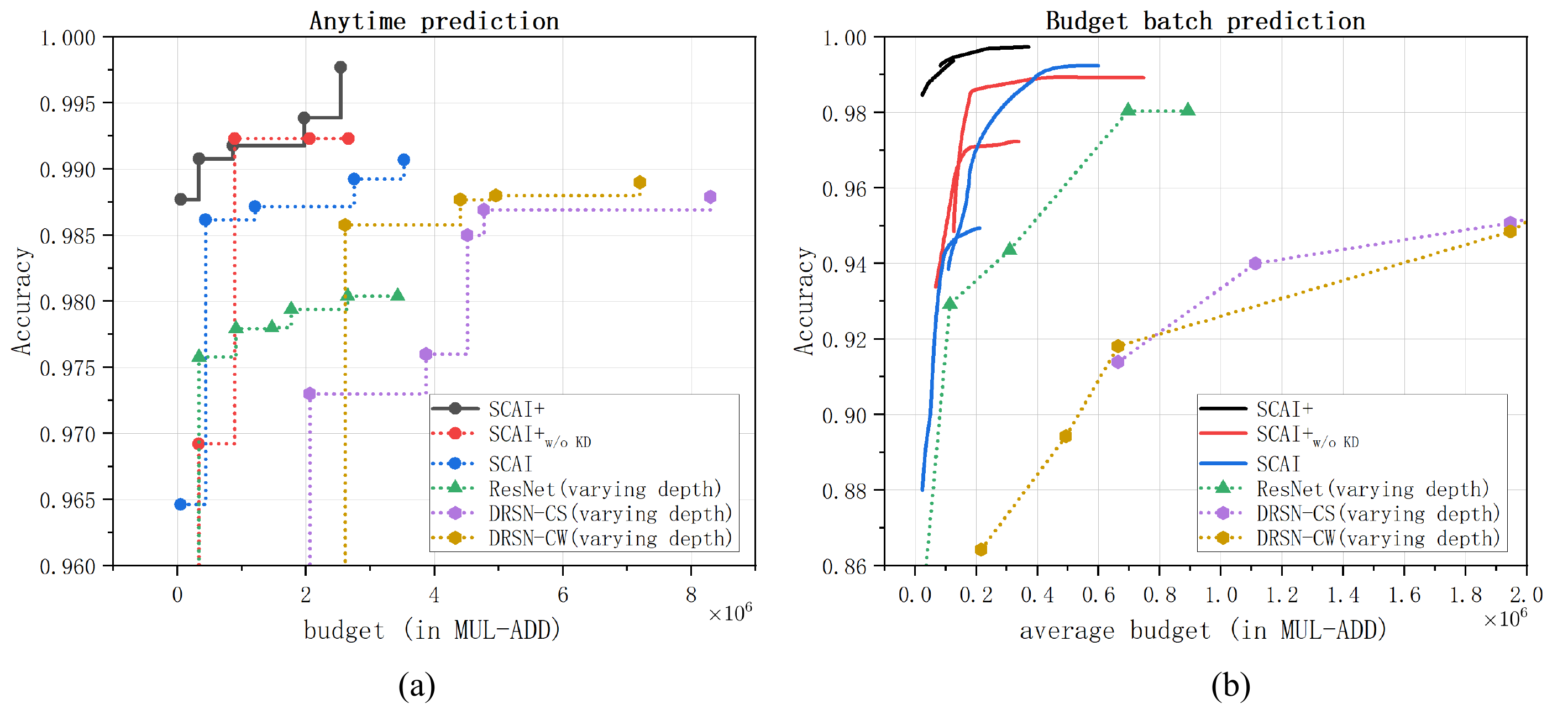}
	\caption{The performance result of different methods with different computational budgets in the (a) anytime prediction and (b) budgeted batch prediction scenario.}
	\label{FIG:9}
\end{figure*}

We compare the above methods in two application scenarios, respectively, and obtain the performance result with different computational budgets as shown in Fig.\ref{FIG:9}. In this section, we will analyze and discuss them in detail.

\subsubsection{Anytime-Prediction Result}\label{sec451}

Fig.\ref{FIG:9} (a) shows the performance result of different methods with different computational budgets in the anytime prediction scenario. It can be seen obviously that SCAI, SCAI+, and $SCAI+_{w/o KD}$ outperform other baseline methods by a large margin. This shows that the early-stop computing architecture proposed in this paper has a significant performance improvement compared to other methods. Let us focus on the left half of Fig.\ref{FIG:9} (a). We can find that SCAI+ and SCAI have more significant advantages than other methods. This phenomenon reflects that the training paradigm of self-distillation can significantly improve the performance of the classifier in the shallow layer of the architecture so that the architecture can achieve higher performance with as fewer computational budget as possible.
As expected, SCAI+ achieves the best performance and makes further performance improvements over other methods under the condition of few computational budgets. This fully verifies our assumption of this paper in section \ref{sec32}: Except for the Raman peaks and fluorescence peaks of different substances with the most informative feature in the spectral curve, the significance in other parts of the curve is relatively small and even is negligible, and most of the curve areas do not need to allocate much computing budget. The reason why SCAI+ achieves a significant performance improvement over other methods under the condition of fewer computational budgets is that the PA-Resnet captures these "class-indiscriminative" features and reduces the computational allocation in these positions. Moreover, the introduction of the self-distillation paradigm improves the performance of the classifier in the shallow layers of the architecture, which also promotes this phenomenon. More interestingly, with the increase of computational budget, the performance of the SCAI+ model reaches a peak that cannot be touched by other models. We believe that the noise reduction effect of the PA-Resnet structure on redundant features makes it focus on "class-discriminative" features and further improves the model performance. The above two points consistently show that SCAI+ can be applied to scenarios with both sufficient and insufficient computational budgets at the same time.

Compared with SCAI, $SCAI+_{w/o KD}$ performs relatively poorly with an insufficient computational budget. We believe that the reason may be that the advantages of PA-Resnet cannot be fully exploited in the shallow model. When the depth of the model increases, PA-Resnet can fully identify important positions of the curve, and selectively allocate the computing budget. Therefore, the performance is also improved rapidly and surpasses that of SCAI, which is consistent with the conclusion of SkipNet \cite{wang2018skipnet}.

Since the threshold units proposed by DRSNs (DRSN-CW and DRSN-CS) cannot learn reasonable thresholds in shallow models, their performance is not comparative with that of ResNet-1D when the computational budget is insufficient. With the increase of computational budget, the threshold unit can correctly capture and filter the redundant or noisy information in the spectral curve, so that the performance of DRSNs is significantly improved. DRSN-CW consistently outperforms DRSN-CS, which is consistent with the original paper of DRSN \cite{zhao2019deep}. This shows that the scheme of applying an independent threshold to each channel of the feature map in the spectral curve works better. Although this scheme needs a relatively larger amount of computation (which can also be ignored), it can better capture noise information, and the performance improvement brought by it can fully compensate for the increase in computation budget.

\subsubsection{Budgeted batch prediction}\label{sec452}

In the Budgeted batch prediction setting, the model accepts a batch of $M$ samples and classifies these samples without exceeding the total computational load budget of $B$. Referring to the experimental scheme of MSDNet \cite{huang2017multi}, we also adopt dynamic evaluation, that is, let "easy" samples be output in the early classifier, and let "hard" samples be output in the latter classifier. For the specific implementation details, please refer to the budgeted batch inference of section \ref{sec331}. To cover a wider range of computational budgets, we train models of SCAI, SCAI+, and $SCAI+_{w/o KD}$ with different scales. The experimental results of this section are reported in Fig.\ref{FIG:9} (b), it can be seen that the experimental results are similar to the results in section \ref{sec451}, SCAI, SCAI+, and $SCAI+_{w/o KD}$ consistently achieve the best performance. For example, SCAI+ achieves approximately 6\% or 13\% accuracy improvement over ResNet-1D and DRSNs (DRSN-CW and DRSN-CS) at an average budget of $0.2\times10^6$ Flops, respectively. To maintain the same accuracy (e.g., 98\%), ResNet-1D requires at least 6 times more computational budget than SCAI, SCAI+ and $SCAI+_{w/o KD}$. This shows that the method proposed in this paper can reasonably allocate computing resources under the premise of a limited total budget, give more computing budget to complex samples, and at the same time, avoid the "over-computation" of simple samples, which is significant for massive data process of computing center in the IoT.

Unlike the results in section \ref{sec451}, ResNet-1D consistently outperforms DRSNs (DRSN-CW and DRSN-CS). This is related to the distribution of the data. Under the current total budget setting, most spectral sample data can obtain the optimal performance of the three models under the condition of few computational budgets. And from the results in section \ref{sec451}, we can see that ResNet-1D performs better in this condition.

\subsection{Computation allocation analysis}

\begin{figure}[t]
	\centering
	\includegraphics[width=3.2in,scale=1.00]{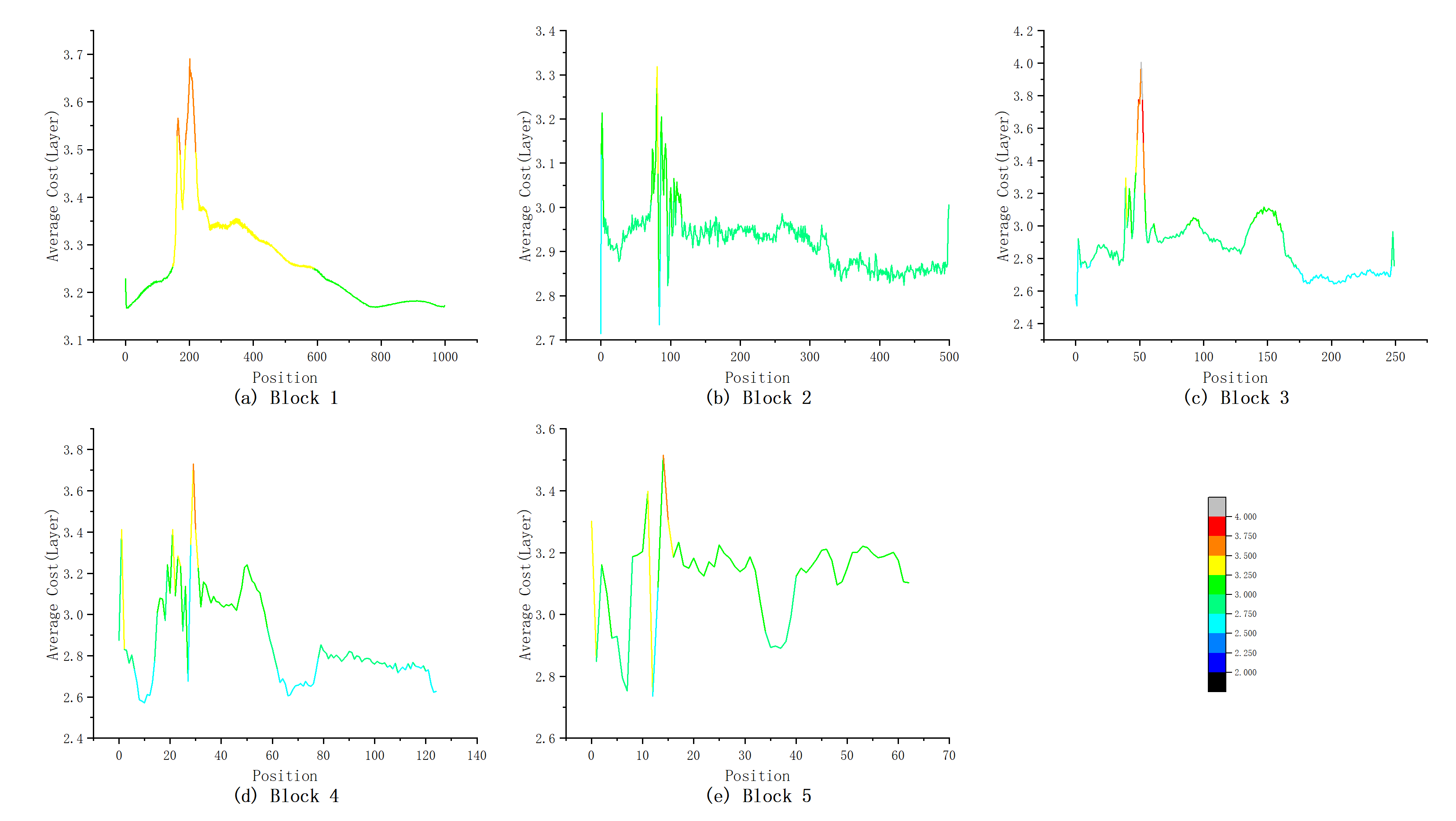}
	\caption{The computation allocation heatmap of PA-Resnet for different curve locations in different blocks.}
	\label{FIG:10}
\end{figure}

A notable feature of our method is the introduction of PA-Resnet. To better explore the role of PA-Resnet and explanation for the improvement of model performance. We plot the computation allocation of the five blocks of the SCAI+ model into the heatmap as shown in Fig.\ref{FIG:10}. From the overall perspective analysis, it can be seen that there are differences in the local positional distribution of the computation allocation of each block. Still, there is a certain similarity in the global distribution. For example, the computation peaks are primarily concentrated in the areas near 200, 100, 50, 25, and 12 positions in the five graphs, respectively. Combined with the analysis in Fig.\ref{FIG:3}, it can be seen that these positions are the Raman peak positions of ethanol, which are "class-discriminative" positions for liquor samples. From the overall distribution of the five graphs, the computing allocation of most locations is much lower than the maximum number of computing layers (4 in this paper), which also explains why the method SCAI+ in this paper can achieve better performance when the computational budget is much lower than that of other methods. This again verifies our assumption of this paper: except for the Raman peaks and fluorescence peaks of different substances with the most informative feature in the spectral curve, the significance in other parts of the curve is relatively small and even is negligible, and most of the curve areas do not need to allocate much computing budget.

\subsection{Self-distillation training analysis}

\begin{figure}[htbp]
	\centering
	\includegraphics[width=3.2in,scale=1.00]{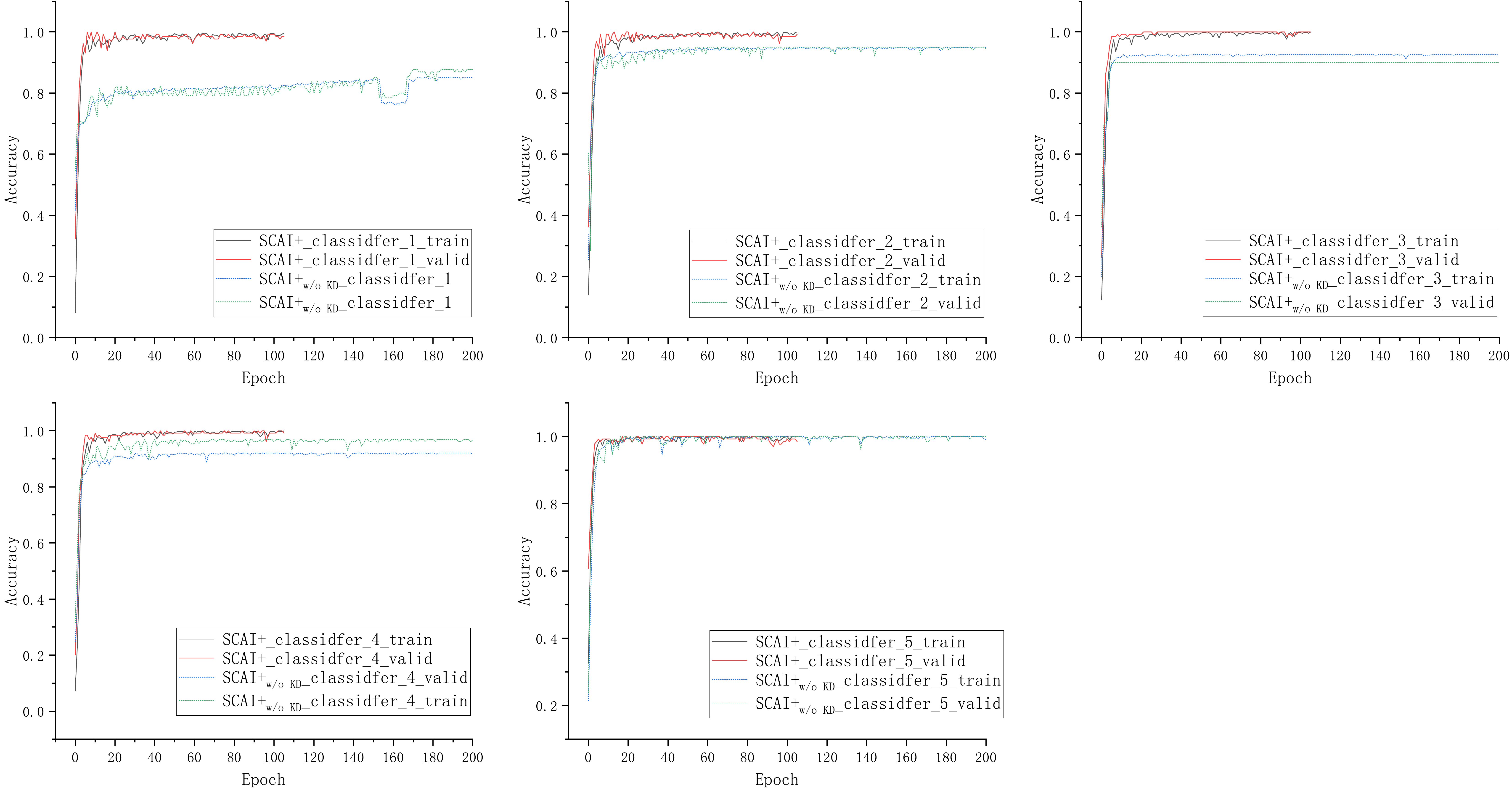}
	\caption{The train and valid performance curves of different classifiers in SCAI+ and $SCAI+_{w/o KD}$.}
	\label{FIG:11}
\end{figure}

Another core module of this paper is the self-distillation training paradigm. We recorded the training data of the classifies within 5 blocks in SCAI+ and $SCAI+_{w/o KD}$, and plotted the train and valid performance curves shown in Fig.\ref{FIG:11}. By comparing the performance of  $SCAI+_{w/o KD}$ and SCAI+ in each figure, we can intuitively see that the two last classifiers achieve roughly the same performance (Figure 11 (e)), but other intermediate classifiers of SCAI+ (Fig.\ref{FIG:11} (a-d)) perform significantly better than that of $SCAI+_{w/o KD}$. More importantly, this advantage is more obvious in the shallower classifier, which converges faster and is more stable after convergence. Taking  Fig.\ref{FIG:11} (a) as an example, SCAI+ converges at least 50 epochs earlier than $SCAI+_{w/o KD}$, and the performance of SCAI+ after convergence is ~18\% higher than that of  $SCAI+_{w/o KD}$. This fully verifies the knowledge guiding role of the last classifier on the previous classifiers in the self-distillation mode, thereby greatly improving the performance of the model, especially under an extremely small computational budget. This can also well explain the phenomenon that the performance advantage of our method in sections \ref{sec451} and \ref{sec452} is more obvious under the condition of less computation allocation. The self-distillation training paradigm plays an important role in extending the application capabilities of this model in the IoT scenario, especially in low-performance smart terminals with high battery life requirements.

\section{Conclusion}

In this paper, aiming at the problem of computational efficiency and computational budget of spectral detection using deep learning models in IoT scenarios, we firstly propose an adaptive inference computing architecture to optimize performance and computational efficiency.

Specifically, to allocate different computations for different spectral curve samples while better exploiting the collaborative computation capability among different devices under the IoT platform, we leverage Early-exit architecture, place intermediate classifiers at different depths of the architecture, and the model outputs the results and stops running when the prediction confidence at the current classifier position reaches a preset threshold. We propose a training paradigm of self-distillation learning, through which the output of the deepest classifier performs soft supervision on the previous classifier to maximize the performance and training speed of the whole architecture, especially the shallow classifier. At the same time, to mitigate the vulnerability of performance to the location and number settings of intermediate classifiers in the Early-exit paradigm, we propose a Position-Adaptive residual network (PA-ResNet). It can adjust the number of layers each block computes at different positions of the curve, so it can not only pay more attention to the information of important positions of the curve (e.g.: Raman peak), but also can accurately allocate the appropriate amount of computational budget based on task performance and computing resources, greatly alleviating the sensitivity of intermediate classifiers settings. The experimental results fully verify the superiority of the method in this paper, especially in the case of limited computing budget, the advantages of the method in this paper are more obvious, which is important for extending its applicability to smart devices( low performance, requirement of long-time battery life or low latency ) in the IoT. The explorational results of the core modules (PA-ResNet and self-distillation training) can also well explain its superiority, which is also in line with our original expectations.

Of course, this paper also has some shortcomings, we need to do some further exploration and research. For instance, more Architecture Design \cite{huang2017multi}, Applicability for More Diverse Tasks \cite{wang2020glance}, and Robustness Against Adversarial Attack \cite{hu2020triple}.


\bibliographystyle{cas-model2-names}

\bibliography{cas-refs}


\bio{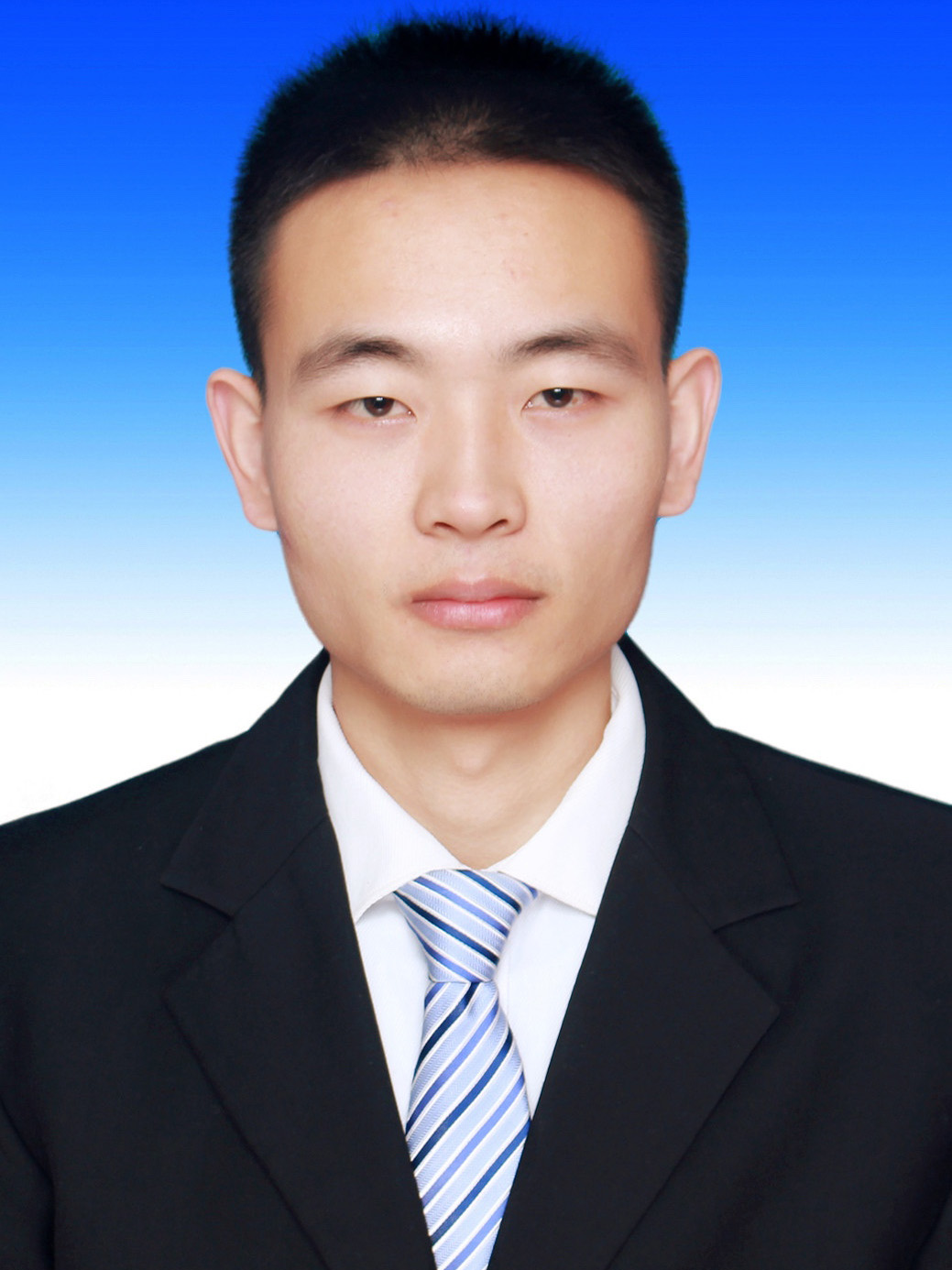}
Yundong Sun is currently working toward the Ph.D. degree in the School of Astronautics at Harbin Institute of Technology. His research interests include machine learning, graph representation learning, recommender system, and spectral data analysis. He has published several articles in journals such as TKDE, IoTJ, Neurocomputing, etc.
\endbio

\vspace{4em}

\bio{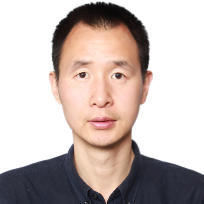}
Dongjie Zhu received the Ph.D. degree in computer architecture from the Harbin Institute of Technology, in 2015. He is an associate professor in the School of Computer Science and Technology at Harbin Institute of Technology, Weihai. His research interests include parallel storage systems, social computing, machine learning, and spectral data analysis. He has published more than 30 articles in several journals or conferences.
\endbio

\vspace{10em}

\bio{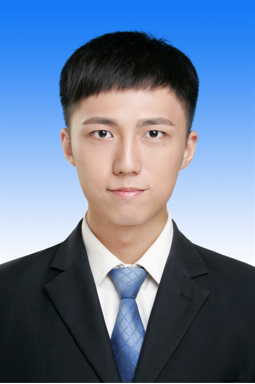}
Haiwen Du is currently working toward the Ph.D. degree in the School of Astronautics at Harbin Institute of Technology. His research interests include storage system architecture and massive data management.
\endbio

\vspace{6em}

\bio{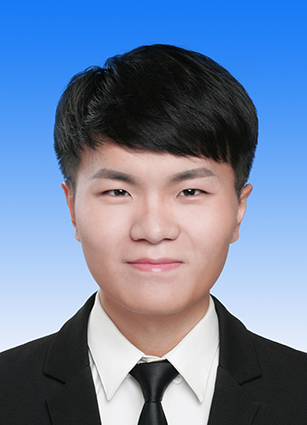}
Yansong Wang is currently working toward a master's degree in the School of Computer Science and Technology at Harbin Institute of Technology. His research interests include machine learning, graph representation learning, recommender system, and spectral data analysis.
\endbio

\vspace{4em}

\bio{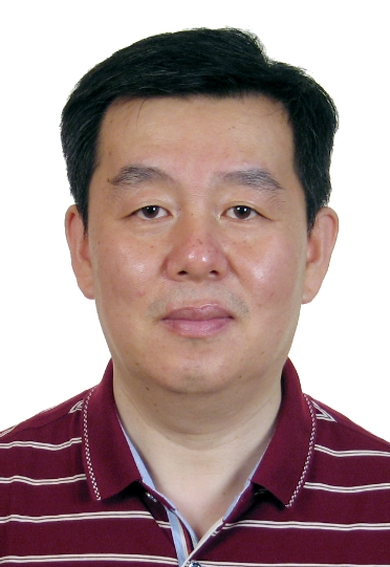}
Zhaoshuo Tian is a professor in the School of Astronautics at Harbin Institute of Technology. His research interests include laser technology and marine laser detection technology. He is a member of IEEE.
\endbio

\end{document}